\def\year{2022}\relax
\documentclass[letterpaper]{article} 
\usepackage{aaai22}  
\usepackage{times}  
\usepackage{helvet}  
\usepackage{courier}  
\usepackage[hyphens]{url}  
\usepackage{graphicx} 
\usepackage{natbib}  
\usepackage{caption} 
\DeclareCaptionStyle{ruled}{labelfont=normalfont,labelsep=colon,strut=off} 
\usepackage[ruled,linesnumbered]{algorithm2e}

\usepackage{float}
\usepackage{multirow}
\usepackage{hyperref}

\usepackage{newfloat}
\usepackage{listings}
\usepackage{enumitem}
\usepackage{mdframed}
\newmdtheoremenv{definition}{Definition}

\usepackage{tikz}

\newcommand{\rules}{\mathrm{VRR}s}

\newcommand{\mask}{\mathrm{mask}}
\newcommand{\where}{\mathit{where}}
\newcommand{\local}{\mathrm{local}}
\newcommand{\sloctzero}{s_{\local}}
\newcommand{\sloctone}{s'_{\local}}
\newcommand{\agentpos}{\mathrm{agent.pos}}

\title{Learning Generalizable Behavior via Visual Rewrite Rules}
\author{
    Yiheng Xie\equalcontrib\textsuperscript{\rm 1},
    Mingxuan Li\equalcontrib\textsuperscript{\rm 2},
    Shangqun Yu\equalcontrib\textsuperscript{\rm 1},
    Michael L. Littman\textsuperscript{\rm 1}
}
\affiliations{
    \textsuperscript{\rm 1}Department of Computer Science, Brown University, \\ 
    \textsuperscript{\rm 2}Department of Computer Science, Columbia University \\
    \{yiheng\_xie, shangqun\_yu, michael\_littman\}@brown.edu, ml@cs.columbia.edu

}

\begin{document}

\maketitle

\begin{abstract}
Though deep reinforcement learning agents have achieved unprecedented success in recent years, their learned policies can be brittle, failing to generalize to even slight modifications of their environments or unfamiliar situations. The black-box nature of the neural network learning dynamics makes it impossible to audit trained deep agents and recover from such failures. In this paper, we propose a novel representation and learning approach to capture environment dynamics without using neural networks. It originates from the observation that, in games designed for people, the effect of an action can often be perceived in the form of local changes in consecutive visual observations. Our algorithm is designed to extract such vision-based changes and condense them into a set of action-dependent descriptive rules, which we call ``visual rewrite rules'' (VRRs). We also present preliminary results from a VRR agent that can explore, expand its rule set, and solve a game via planning with its learned VRR world model. In several classical games, our non-deep agent demonstrates superior performance, extreme sample efficiency, and robust generalization ability compared with several mainstream deep agents.
\end{abstract}

\section{Introduction}
While deep reinforcement-learning agents have achieved impressive performance in Atari games~\citep{impala, agent57, muzero, dreamerv2}, the vast amount of data they require for training and their limited generalization ability preclude us from extending these approaches to more meaningful and challenging real world tasks. The sample efficiency problem has been long existed in the reinforcement-learning (RL) literature due to the challenges of the ``curse of dimensionality''~\citep{dp}. Even deep RL agents must contend with the hardness of learning good representations~\citep{trainDL, rad}. For the generalization issue, studies~\citep{procgen, coinrun, geneindrl} have demonstrated that deep RL agents can easily memorize and overfit to training environments instead of truly understanding the dynamics that underpin the task. As a consequence, even simple variations in an environment that human learners barely even notice can undermine the performance of a deep RL agent.

Though there are extensive studies trying to address the aforementioned problems~\citep{DBLP:conf/iclr/LeeLSL20, procgen, rad}, most continue to adopt an end-to-end learning framework. We show that by disentangling representation learning from policy learning, problems can successfully be solved in parts.

Specifically, human game playing priors~\citep{gameprior,tsividis2017human} support the learning of high-level factored rules built upon the concept of an object~\citep{oomdp}, which support planning in a wide variety of related tasks. Motivated by the observation that it is possible to reason~\cite{furnas91} and compute~\cite{ackley18} with picture-to-picture rules, we examine using such an approach to learn environmental dynamics in RL. 
We present \emph{visual rewrite learning}, a simple learning algorithm that captures local action-dependent visual rewrite rules (VRRs) of the environment. We also present a model-based agent framework integrating VRR and planning. 

We demonstrate the effectiveness of the VRR agent in a series of 
publicly available environments. Our VRR agent outperforms state-of-the-art model-free and model-based deep RL agents in tests of generalization and uses significantly fewer training samples. To the best of our knowledge, we are the first to beat deep networks with a non-deep approach in these environments.


\begin{figure*}[ht]
  \centering
  \includegraphics[width=\linewidth]{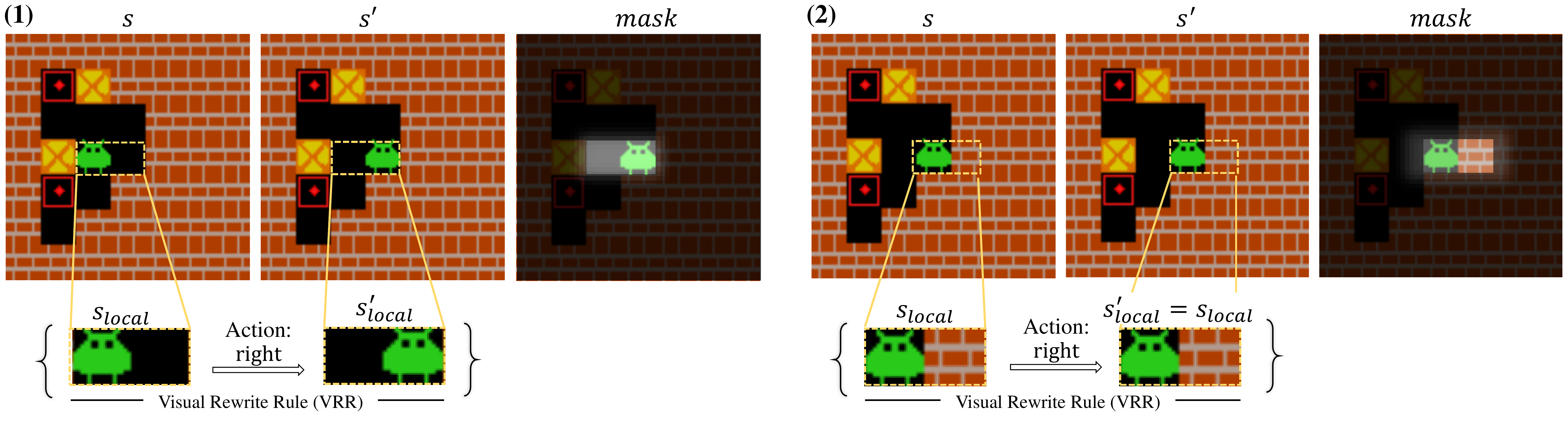}
  \caption{Example Visual Rewrite Rules (VRRs) in Sokoban. State change can be decomposed into a static non-local component and a dynamic local component near the agent. The local region is highlighted by $\mask$. (1) The Sokoban agent (green) moves right onto the empty space, in response to action $\mathrm{right}$. (2) Agent cannot move into the brick wall with action $\mathrm{right}$, instead no state change occurs ($s^t = s^{t+1}$).}
  \label{fig:vrr_illustration}
\end{figure*}


\section{Background}

In this work, we use Markov decision processes (MDPs) to model an agent interacting with its environment while solving a task. An MDP is defined by the tuple $\langle \mathcal{S}, \mathcal{A}, \mathcal{R}, \mathcal{P}, \gamma\rangle$ where state $s \in \mathcal{S}$, action $a \in \mathcal{A}$, reward $r\in \mathcal{R}$, state transition described by $\mathcal{P}(s_{t+1}, r_t|s_t, a_t)$ and discount factor $\gamma \in [0, 1)$. The RL agent's goal is to maximize discounted reward $\sum_{t=0}^\infty \gamma^t r_t$. 

In model-based reinforcement learning, an agent learns an environment model composed of a state-transition model and a reward model. It then uses the learned model to navigate through the environment, gathering reward. Dyna~\citep{dyna}, which integrates model-free learning with generated data from a learned model, is an example of such an approach. 
Deep RL, with its focus on visual observations, was slow to incorporate model-based learning. But, by now, 
deep world models are being used to generate agent-simulated trajectories for training model-free agents~\citep{wm, iaa, metamodel, simple, dreamerv2} or can be integrated with lookahead planning~\citep{muzero, genegame, roleplanning}. A major challenge is that the quality of predicted trajectories typically degrades quickly as small errors compound as trajectories are rolled out~\citep{mbpo,asadi19}. In addition, overfitting hampers the application of a deep-learned world model to predictions on out-of-distribution (training) states, even when they share dynamics. We argue that without an organized, structural way of learning and using the learned environment dynamics, it is nearly impossible to verify if the world model has grasped the essential knowledge for solving the task and therefore applying to novel situations. We investigate a more structured non-deep-network-based approach for learning a world model.

Our approach can be viewed as a form of abstraction~\citep{li06, neceabs} in which observations are decomposed into patches of pixels observed to change together. Two main forms of abstraction in RL are state abstraction and action abstraction. The former maps the original larger state space into a smaller state space while preserving the essential properties for solving the task~\citep{DBLP:conf/aaai/DeanG97, DBLP:conf/ijcai/JongS05, DBLP:conf/icml/JiangKS15, DBLP:conf/icml/AbelALL18}. The latter approach aggregates the atomic actions into new units enabling high level planning and skill reuse~\citep{DBLP:journals/ai/SuttonPS99, DBLP:conf/nips/KonidarisB09, DBLP:conf/iclr/SharmaGLKH20}.


\section{Visual Rewrite Rules}
We introduce Visual Rewrite Rules (VRRs) as a new representation for describing game-environment dynamics. In contrast to deep neural networks, VRRs are an intuitive, elegant, yet robust method to model environments. In this section, we first offer an intuitive explanation of how VRRs describe state transitions. Then, we introduce the formal definition of visual rewrite rules, along with their strengths and limitations. We then explain how to learn VRRs from experience data. Finally, we present subtleties to implementing VRRs with pure-vision observations. 
\begin{figure}[h]
  \centering
  \includegraphics[width=\linewidth]{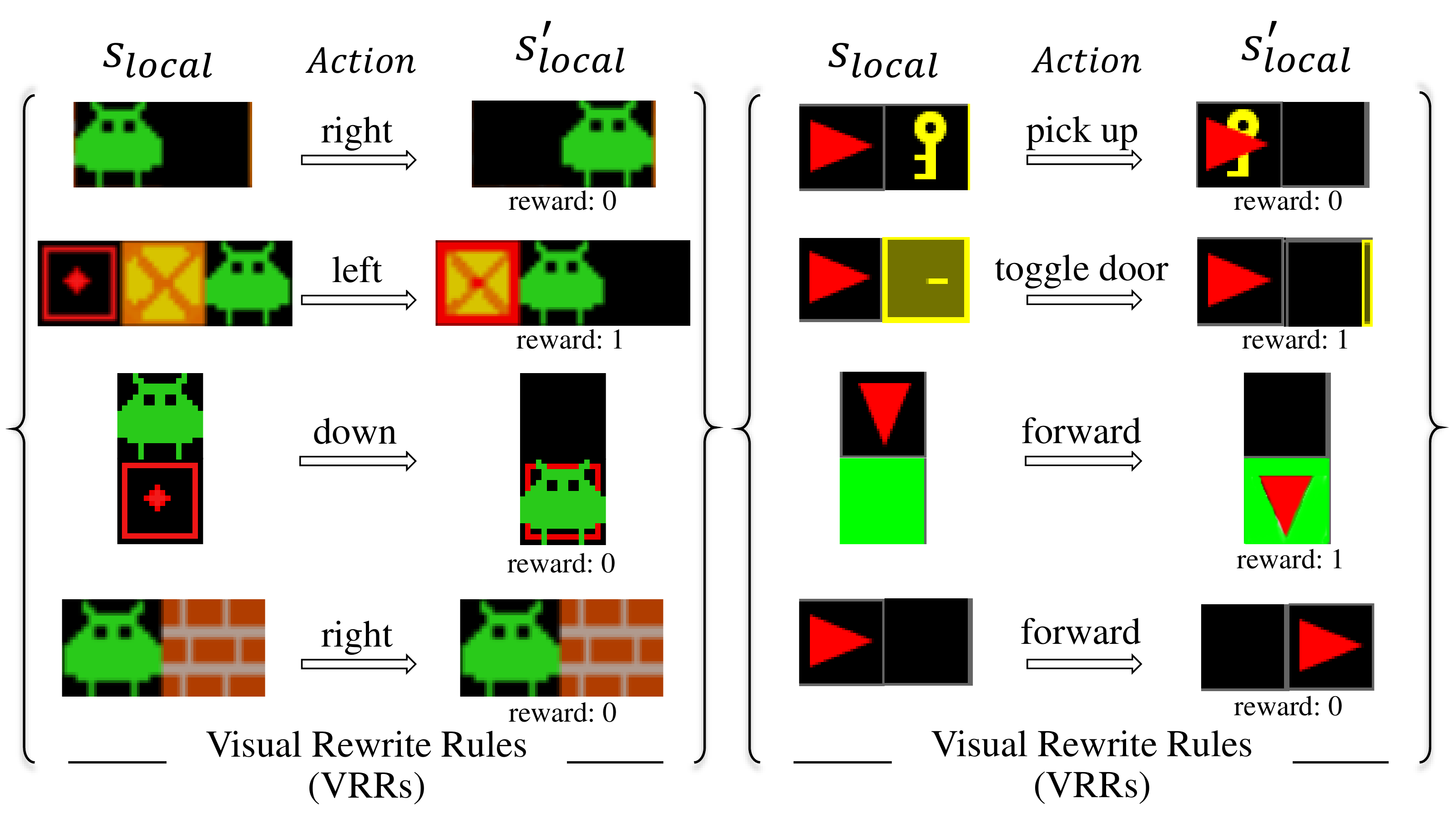}
  \caption{Example Visual Rewrite Rules in Sokoban (left) and MiniGrid (right). }
  \label{fig:vrr_dict}
\end{figure}

\subsection{State Transitions as VRRs} 

Intuitively, one may think of a VRR state transition as updating a subset of the state vector representing the state according to the action, while the rest of the state vector remains the same. The dynamic, changing component is ``rewritten" based on the outcome of the action, hence the name ``visual rewrite''. 
Fig.~\ref{fig:vrr_illustration} shows an example of modeling the state transitions in Sokoban using VRRs. 
In the first graph of Fig.~\ref{fig:vrr_illustration}, the agent moves from one blank cell to another blank cell under action ``move right". All other grid cells remain the same between the two time steps. The two cells are ``rewritten" with the outcome of the action. 
In the second graph of Fig.~\ref{fig:vrr_illustration}, the agent tries to move right but bumps into a wall. There is no state change. 
The reason that the same action results in different transitions is explained by the two cells located around the agent, the difference being the presence and absence of a wall.

In many games, the effects of agent actions is localized near the agent. 
Hence, VRRs decompose the state transition into a \textit{local} component near the agent, and a static non-local component. 
We use \textbf{Visual Rewrite Rules (VRRs)} to model such state transitions. 
Each VRR describes the shape of the local component of the state vector, what the local component looks like, how the local component changes under the influence of actions, and the reward. Fig~\ref{fig:vrr_dict} shows a selection of VRRs from the Sokoban and MiniGrid games. 

Formally, each VRR is defined as $\mathcal{F}: (s[\mask], a) \rightarrow (s'[\mask], \mathrm{reward})$ mapping the current local component of $s$ and action $a$ to the local component after the transition and reward. We define $\mask$ as an indicator function for the shape of the local component of the state vector, specifically $s_{local} = s[\mask]$. 

When a state transition occurs (as in Figure~\ref{fig:vrr_illustration}, left), $\mask$ indicates the position where the pixel values have changed between $s$ and $s'$: $\mask = \where(s \neq s')$. 
The set of VRRs are stored as a \textbf{dictionary}, where the key-value pair is $\left((s[\mask], a), (s'[\mask], \mathrm{reward})\right)$.

\subsection{VRR World Model}

VRRs form a vision-based world model. At inference time, a VRR can be ``stamped'' onto a new state vector to predict state transitions.
Algo~\ref{alg:vrrworld} describes the VRR world model.

\begin{algorithm}[h]
	\caption{\textit{world\_model}. VRRs as a world model.} 
    \label{alg:vrrworld}
    \SetKwInOut{Input}{Input}
    \SetKwInOut{Output}{Output}
    \Input{State vector $s$. Action $a$. Rule set $\rules$. Agent position $\agentpos$.}
    \Output{State vector $s'$. $\mathrm{reward}$. status.}
    $\mask = \where(s \neq s')$\;
    \tcp{Compute local component.} 
    $\sloctzero = s[\mask]$\;
	\For {$s_{rule} \in \rules[a]$} {
        $tmpS_{rule} = s_{rule}$ shifted to $\mathrm{agent.pos}$\;
        Compare $tmpS_{rule}$ with current $\sloctzero$\;
        \tcp{Found a match.}
        \If {$tmpS_{rule} = \sloctzero$} {
            $(\sloctone, \mathrm{reward}) = \rules[a, s_{rule}]$\;
            Update $s$ with $\sloctone$ to obtain $s'$\;
            $\mathrm{status} = $ known rule\;
            return $s', \mathrm{reward}, \mathrm{status}$\;
    	}
    }
    \tcp{No matching local states found.}
    $\mathrm{status} = $ unknown rule\;
    return $\mathrm{status}$\;
    
\end{algorithm}
Given a state vector $s$ and agent position $\agentpos$, we compare the local region centered around the agent $s_{\local}$ to each local component $s_{rule}$ stored in the $\rules$ dictionary keys. 
If there is a match, we use the tuple of agent action $a$ and $s_{\local}$ to find the resulting state transition $\sloctone$, and reward $r$. 
If there's no matching dictionary entry, this indicates a previously unseen local component of the state vector.

By the nature of most games, the number of VRRs needed for constructing a perfect VRR world model is small, since the game dynamics is usually composed of basic components such as movements, pick up, and toggle. We leverage this sparsity of game rules to learn an efficient set of VRRs. 
VRR is also invariant to rotation and translation, since we only focus on the local action effect while ignoring the static components. This leads to robust and generalizable world model, which we show in the experiments section.

\subsection{Learning VRRs}
From the definition of VRR, the most straightforward way of learning it is by contrasting $s$ with $s'$, and recording the pixels that are affected by the action. Algo~\ref{alg:vrr_basic} illustrates this basic idea of learning VRRs.
\begin{algorithm}
	\caption{Learning VRR from state difference.}
    \label{alg:vrr_basic}
    \SetKwInOut{Input}{Input}
    \SetKwInOut{Output}{Output}
    \Input{Game state vector (grid) from two adjacent time steps $s$, $s'$. Action $a$. Global Rules set $\rules$}
    \tcp{Mask out the unchanged region.}
	$\mask = (s \neq s')$\;
	\tcp{Store the change in dictionary.}
	$\rules[a, s[\mask]] = s'[\mask]$\;
\end{algorithm}

But there is subtlety here. Normally, one would expect that every action leads to some changes in the visual representation of the environment. However, it is very common that certain state--action pairs result in no state change at all. For example, in Sokoban, if the agent tries to push the box into a wall, $s'$ is exactly the same as $s$ (Fig.~\ref{fig:vrr_illustration} part 2).
In such cases, $s=s'$ leads to an empty mask. That is, $\mask = where(s \neq s') = \emptyset$, $\sloctone = \sloctzero = \emptyset$.

A naive solution is to store the entire game state, but this is impractical, and defeats our goal of building a compact, canonical rule set.
Therefore, an appropriate inductive bias is necessary for choosing \emph{where} to look for evidence that could explain the absence of state change (for example, the wall in front of the agent).

We extend Algo~\ref{alg:vrr_basic} into Algo~\ref{alg:exp_corr}, where the ``if'' statement explains why a state transition occurred, while the ``else'' clause seeks to explain why a state didn't change under certain actions. In the latter case, the \textit{local} region is where the VRRs \textit{expect} changes to happen. 

For example, a previously learned VRR expects the agent to move forward with action ``forward", where the input of this VRR has a two-cell local component---the agent, and the empty space in front of it. When such state transition does not occur, we first mask out the same local component and notice that the empty space in $s_{rule}$ is now a wall, which explains the absence of change. In the case when no state change is ever observed with a certain action, VRRs simply assume this action does not lead to state changes. When this is proven to be wrong, VRR will record the local component change, and proceed as normal.

With this strategy, VRR can now handle the situation when null state transition happens. 
\begin{algorithm}[ht]
	\caption{\textit{learn\_vrr}. Learning VRRs with expectation correction.} 
    \label{alg:exp_corr}
    \SetKwInOut{Input}{Input}
    \SetKwInOut{Output}{Output}
    \Input{State vectors $s$, $s'$. $\mathrm{reward}$. Action $a$. Rules set $\rules$. Agent position $\agentpos$.}
    \Output{Updated rules set $\rules$.}
	$\mask = (s \neq s')$\;
	\eIf {$\mask \neq \emptyset$} {
        $\rules[a, s[\mask]] = (s'[\mask], \mathrm{reward})$\;
	}{
        \tcp{Given no state change occurred, we take the existing VRRs' local scope as $\mask$.}
        $\mask = $ intersection of all $\sloctzero$ from current $\rules$\;
        $\mask' = \mask$ shifted to $\agentpos$\;
        $\rules[a, s[\mask']] = (s[\mask'], \mathrm{reward})$\;
    }
\end{algorithm}

\subsection{VRR in Practice} 
\subsubsection{Agent Position}
In previous sections, we assume that the agent is not only given the observation but also knows its own position. In other words, the controllable object, which is the agent, is differentiated from other game objects. This information is crucial because the locality of visual changes is defined relative to the agent's position. Though identifying the controllable object seems to be intuitive for us, it's not so obvious for an algorithm without any prior knowledge that human players do. 

To relax this assumption, we use a principled method to identify the agent in the game. We define the agent as the object that exhibits the ability to independently move as a result of input actions. Other passive game objects, such as boxes in Sokoban, can only move as a result of the agent's actions. Hence, state changes always occur near the agent, since the object exhibiting agency is always present to produce those state changes. We leave handling more complex games with independent moving objects (Frogger, Space Invaders, etc.) for future work.

\subsubsection{Object Centric Representation}
\begin{figure}[b]
  \centering
  \includegraphics[width=\linewidth]{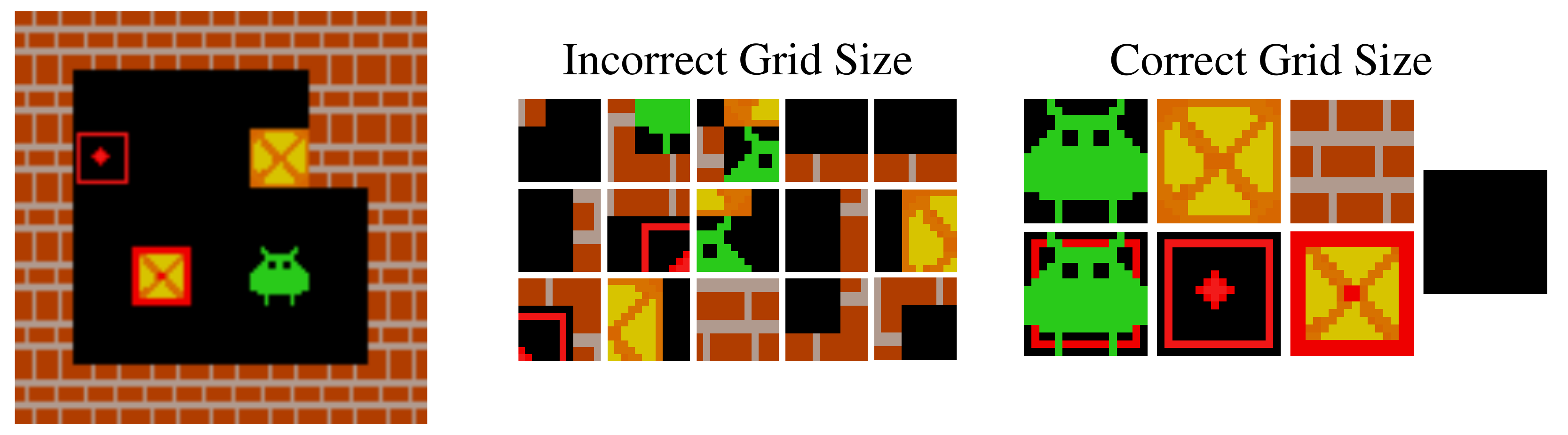}
  \caption{Correct grid size results in a minimal number of object types, while incorrect grid slicing results in a combinatorial number of object types.}
  \label{fig:slice}
\end{figure}

During implementation, we adopt a discrete-space and object-centric representation of the state space. Given a rendered game observation in pixels, we discretize the screen into squared grid cells. The grid size is \textit{not} assumed to be given. Instead, we search for the minimal sprite unit by picking the grid size that yields the least number of object types while also being semantically meaningful (Fig.~\ref{fig:slice}). An erroneous grid size (for example slicing between the ground truth grids) would result in an explosion of the number of object types and the number of rules. At last each distinct object type is assigned with an unique id. This approach transforms the visual observation into a compact yet expressive object-oriented state vector that also preserves the spatial topology. 


Although we only show results for games whose observation space is discrete and two-dimensional, in principle, VRRs are compatible with games that do not fit into a regular grid, such as balls and paddles in Pong. 
Additionally, the idea of VRR can also model stochastic state transitions easily by extending the dictionary keys to a \textit{distribution} of possible future states. We reserve these topics for future work.

\section{A VRR Agent}

\begin{algorithm}[ht]
	\caption{VRR agent.} 
    \label{alg:vrr_agent}
    \SetKwInOut{Input}{Input}
    \SetKwInOut{Output}{Output}
    \Input{Game environment $\mathrm{env}$. Rules set $\rules$.}
    \While {not done} {
        BFS from current state, using VRR world model\; \tcp{Algo.~\ref{alg:vrrworld}}
        $\mathrm{BFStree} = $ empty tree, where nodes are states, edges are actions\; 
        \tcp{BFS keeps track of reward, and $\mathrm{status}$}
        \uIf {BFS maximum reward $> 0$} {
            $\mathrm{actions} = $action sequence from root to winning state\;     
        } \uElseIf {$\mathrm{status}$ = new rule} {
            \tcc{If agent cannot complete the game with current $\rules$, explore new rules.}
            $\mathrm{actions} = $action sequence from root to novel state\;
        } \uElse {
            \tcc{No more new rules, games not winnable (such as a box in corner in Sokoban).}
            agent gives up this round\;
        }
        \For {$a \in \mathrm{actions}$} {
            $s, s', \mathrm{reward} = \mathrm{env}.step(a)$\;
            \tcp{Expand $\rules$ with Algo.\ref{alg:exp_corr}}
            $\rules = learn\_vrr(s, s', \mathrm{reward}, a, \rules, \agentpos)$\;
        }
    }
\end{algorithm}

A VRR agent is composed of a VRR world model described above and a compatible planning algorithm. We choose breadth-first search (BFS) for its simplicity, but VRR is trivially compatible with more sophisticated search algorithms such as Monte-Carlo Tree Search~\cite{mcts}. 

To learn the VRRs efficiently, a principled method for exploring the game is needed. As described in Algo~\ref{alg:vrr_agent}, the agent first attempts to solve the game via BFS planning with the current set of learned VRRs.
If BFS returns an action sequence that leads to the winning state, then the set of learned game rules is sufficient for the agent to solve the task. The agent then executes this action sequence to receive the reward and moves onto the next round without wasting any time. Conversely, if BFS terminates without finding the winning state, there must exist state transitions that the agent is yet to learn, that is, $\exists\, \mathrm{game\, rule} \notin \rules$. 

\begin{figure*}[ht]
  \centering
  \includegraphics[width=\linewidth]{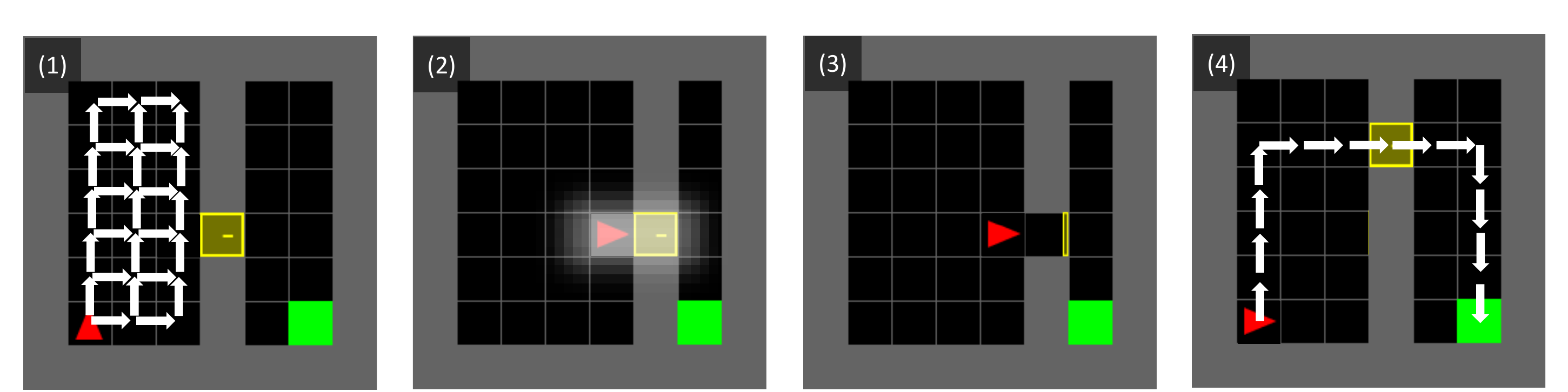}
  \caption{VRR agent explores new rules when its current knowledge is insufficient to solve the game. It balances exploration and exploitation leading to sample-efficient learning. For a live demo, see \href{https://youtu.be/P7USC23NwXs}{Sokoban experiments} and \href{https://youtu.be/OmqycU3f2UU}{DoorKey experiments}.}
  \label{fig:interpretable}
\end{figure*}

The VRR world model is capable of indicating when it has encountered an unknown local component rewrite. That is, when a $\mathrm{key}$ is absent from the $\rules$ dictionary. In this circumstance, BFS planning will return an action sequence that guides the agent to explore this new transition.
Fig.~\ref{fig:interpretable} illustrates such an example. If the agent has not yet interacted with the door, and thus cannot solve the game, the BFS will return action sequences that explore unknown state transitions. One of such action sequence will navigate the agent to the door, and then execute the action ``toggle.'' The new rule describing door-opening will then be recorded in $\rules$.

A notable exception here is Sokoban, in which irreversible behaviours exist. When agent pushes a box into a corner, it can never get that box out. VRR agent is able to identify such situations: neither is the game solvable, nor are there any unknown states to be explored. The agent will promptly give up the current round.



A VRR agent can learn game rules without any prior knowledge from scratch with extreme sample efficiency. At test time, VRR agent is capable of zero-shot generalization to new game levels, which we demonstrate in experiments. VRR agent learning is also amenable to lifelong learning, where it can learn a subset of game rules and update its knowledge in an online fashion as the game complexity increases. Moreover, devoid of any black box components, the VRR agent's learning and planning is completely interpretable, as shown in Fig.~\ref{fig:interpretable}.

\section{Experiments}
\label{sec:exp}

\begin{figure}[ht!]
  \centering
  \includegraphics[width=\linewidth]{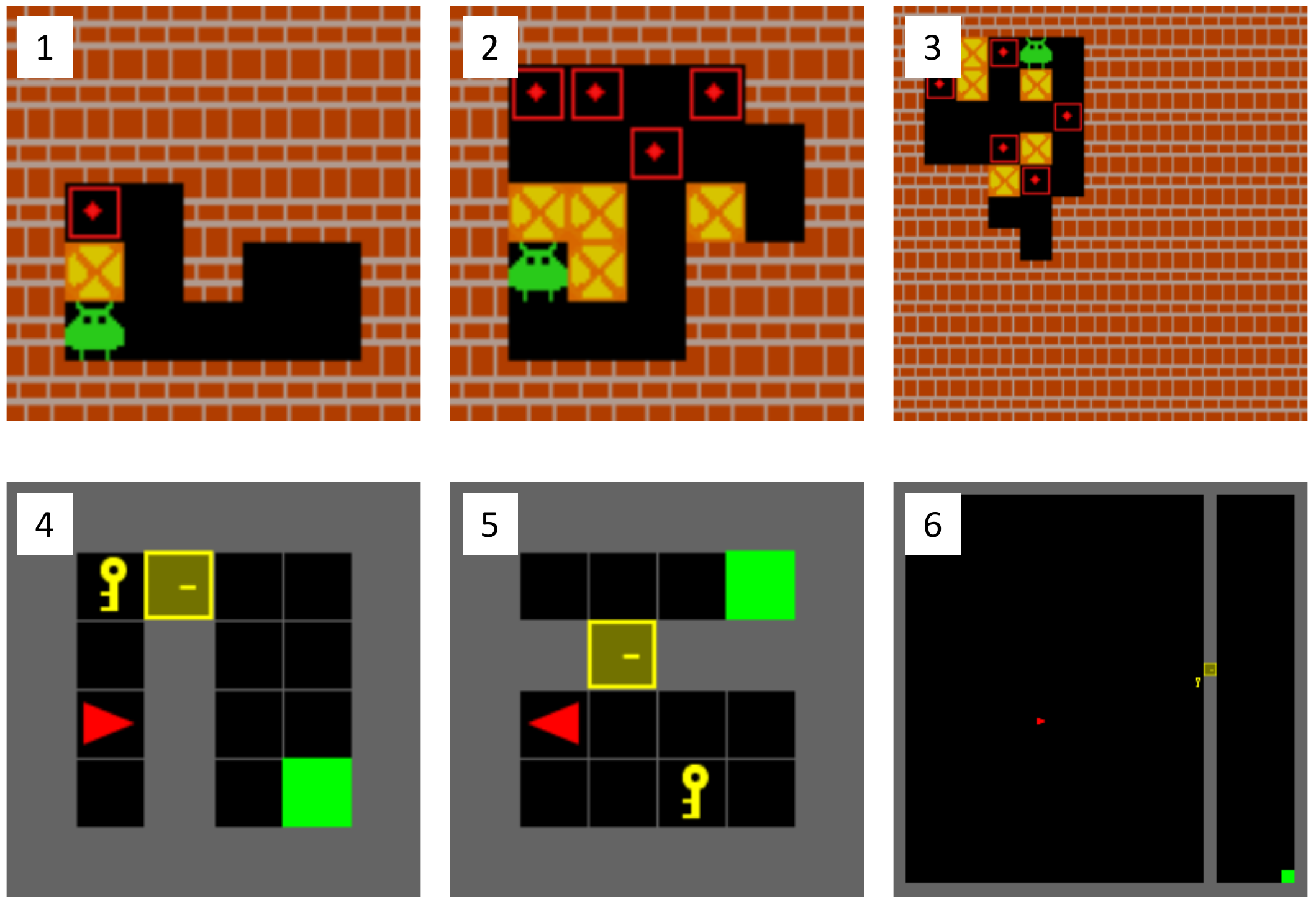}
  \caption{Game environments. 1) $7\times 7$ Sokoban with 1 box (train), 2) $7\times 7$ Sokoban with 4 boxes (test), 3) $13 \times 13$ Sokoban with 5 boxes (test), 4) $6\times 6$ DoorKey (train), 5) $6\times 6$ DoorKey rotated (test), 6) $32\times 32$ DoorKey (test).}
  \label{fig:game_desc}
\end{figure}

In this section, we demonstrate the sample efficiency and generalization ability of our VRR agent in comparison with several deep RL agents: PPO~\cite{ppo}, IMPALA~\cite{impala}, and DreamerV2~\cite{dreamerv2}. With the exception of using the original implementation for DreamerV2, we adopt the Ray RLib~\cite{ray} version for our baselines. 
The following environments are used for training and evaluation: MiniGrid~\cite{gym_minigrid} and gym-Sokoban environment~\cite{gymsokoban}. They are both procedurally-generated environments with varying game layouts and clearly-defined difficulty levels. See Fig.~\ref{fig:game_desc} for a brief description of the training environments and their variations. When reporting the average return of deep RL baselines, we show the mean and standard derivation from 3 independent runs. The average return is calculated as the rolling mean of the last 10 episodes. We will release our source code soon.

\begin{figure*}
  \centering
  \includegraphics[width=\linewidth]{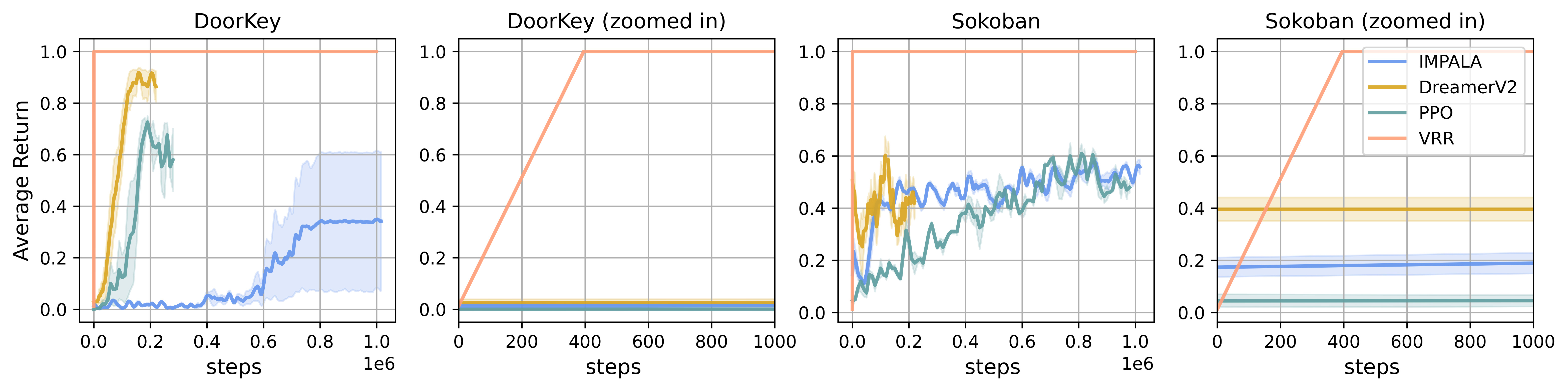}
  \caption{Average return during training. From left to right: DoorKey ($6\times 6$), DoorKey ($6\times 6$) zoomed, Sokoban ($7\times 7$), Sokoban ($7\times 7$) zoomed. The VRR agent achieves the same reward while using 3 orders of magnitude fewer training steps.}
  \label{fig:exp_train}
\end{figure*}

\subsection{Sample Efficiency in Training} 

\begin{figure}[H]
  \centering
  \includegraphics[width=\linewidth]{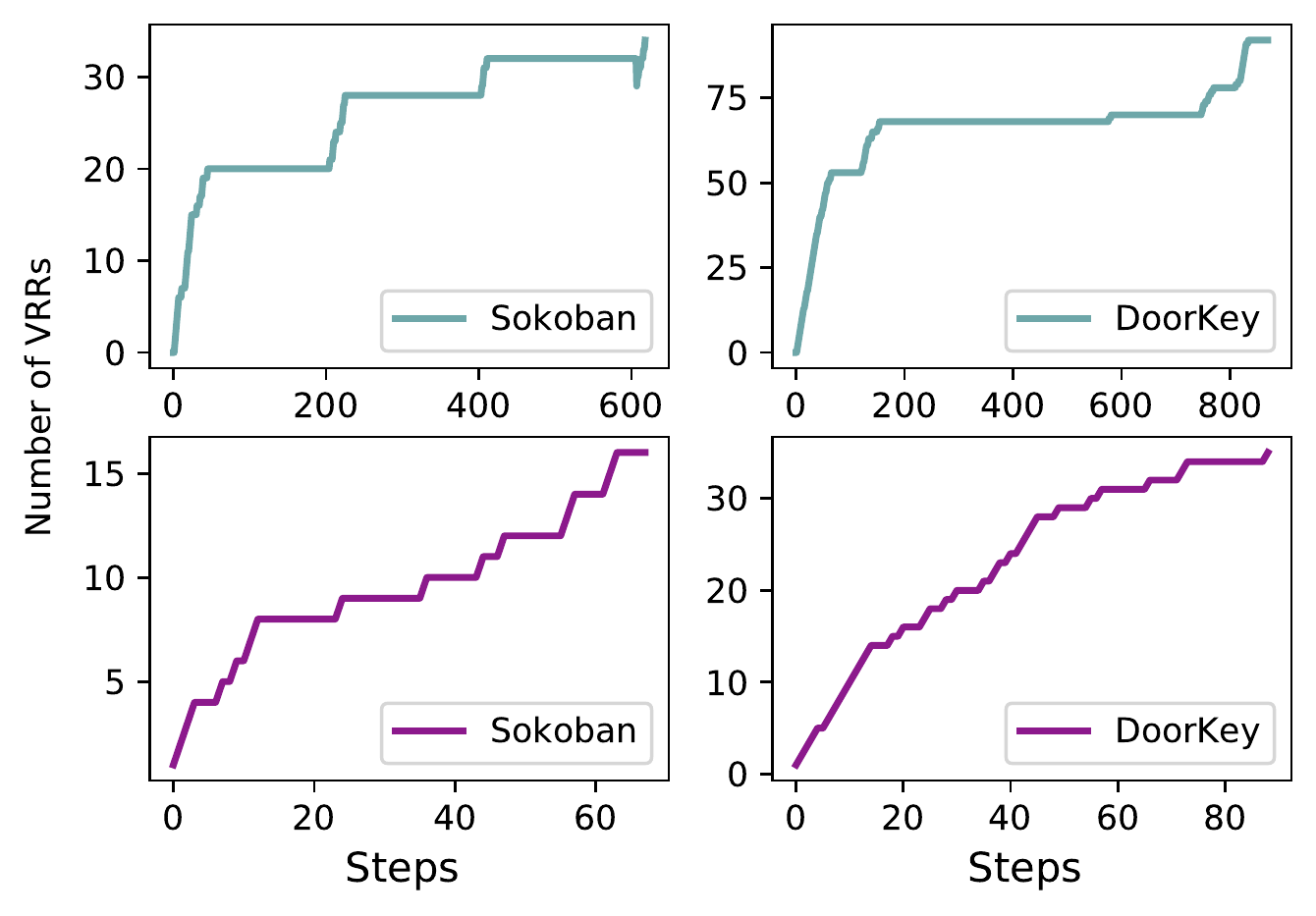}
  \caption{Number of VRRs learned during training as a function of training steps. Top: learning from scratch by interacting with the game environment. Bottom: learning from an extremely small set of human play data.}
  \label{fig:train_num_rules}
  \label{fig:exp_guided}
\end{figure}

As described in Algo~\ref{alg:vrr_agent}, we train the VRR agent from scratch with an initially empty rule set. By interacting with the game environment, the VRR agent gradually expands its rule set (as shown in Fig.~\ref{fig:train_num_rules}, top). 
In Fig.~\ref{fig:exp_train}, we show that the VRR agent requires orders of magnitude fewer environment steps than the deep baselines before converging. In Sokoban, the VRR agent also achieves higher return at convergence. 

\begin{table}[H]
\centering
\begin{tabular}{c c c c c}
\hline
Game                                                & Data Src.    & Steps & Avg. Return & Avg. Steps \\
\hline
\multirow{2}{0.15\columnwidth}{Sokoban} & scratch      & 440       & 1.0   &  4.528 \\
                                                      & human        & 81        & 0.96  &  4.239 \\
\hline
\multirow{2}{0.15\columnwidth}{DoorKey} & scratch      & 1058      & 1.0   &  12.724  \\
                                                      & human        & 89        & 1.0   &  10.249 \\
\hline
\end{tabular}
\caption{VRR agent trained on a small human play dataset achieves comparable performance at test time, while using an order of magnitude fewer training steps. Sokoban and DoorKey game board sizes are  $7\times 7$, and  $6\times 6$, respectively.}
\label{tab:efficiency}
\end{table}

\subsubsection{Minimal Guided Learning}
The sample efficiency of the VRR agent is explained by that the rule set captures only the novel and necessary local component transitions without redundancy. However, the sample efficiency is ultimately bounded by the exploration efficiency of the agent. The faster the agent discovers new game rules, the sooner it can solve the game. 

To test the VRR agent's sample efficiency in the limit. 
We demonstrate that VRR agent can learn from an extremely small training set of under 100 steps (Fig.~\ref{fig:exp_guided}), while achieving identical performance as learning from scratch (Table~\ref{tab:efficiency}). The dataset is collected from a human player, who solves a few levels of the game while demonstrating all the basic movements required to complete the task. Such a limited dataset is completely insufficient for training deep RL agents, for whom the required number of training steps before convergence is often 4-5 orders of magnitude larger. This resonates with our argument that by disentangling representation learning from policy learning, the sample efficiency problem will be greatly alleviated. So does the generalization problem, as we show below.

\subsection{Generalization Experiments}
Next, we test the zero-shot generalization performance of VRR agent. Since both DoorKey and Sokoban are procedurally generated, their underlying game rules remain unchanged regardless of game parameters. 
The game parameters we investigate include (Fig.~\ref{fig:game_desc}):
\begin{itemize}
    \item number of boxes in Sokoban,
    \item random rotation to initial states in DoorKey,
    \item game board size (such as $7\times 7$ vs.\ $13\times 13$).
\end{itemize}
We show that the VRR  agent can solve more complex game levels with the essential rule set learned from the most basic level. 
For all zero-shot generalization experiments below, we train VRR and baseline agents on $7\times7$ Sokoban environment with 1 box, and $6\times6$ DoorKey environment.

\begin{table}[b]
\centering
\begin{tabular}{c c c c c}
\hline
                    & \multicolumn{4}{c}{Average Return}\\
Game                & VRR & Dream & IMPALA & PPO\\
\hline
Sokoban ($7\times7$)     & 1.0   & 0.65 & 0.76 &   0.64         \\
Sokoban ($13\times13$)   & 1.0   & 0.04	& 0.11 &   0.0      \\
\hline
DoorKey ($6\times6$)     & 1.0   & 1.0  & 1.0  &   1.0      \\
DoorKey ($32\times32$)   & 1.0   & 0.0  & 0.0  &   0.0      \\
\hline
\end{tabular}
\caption{\label{tab:grid_size} Zero-shot performance with varying grid size. Note: Sokoban is the 1-box map. Dream: DreamerV2.}
\end{table}
\subsubsection{Sokoban}
There are primarily two game parameters that affect the complexity of the game: number of boxes and board size. 
We test the performance of VRR agent in larger maps with the same number of boxes used during training. 
Table~\ref{tab:grid_size} shows that VRR agent performance is invariant to Sokoban board size, while baselines struggle to generalize to larger board sizes. 
Empirically, a larger map enables more complex layouts and larger search spaces, which makes it easier to trap the agent in irreversible situations.
Furthermore, CNN-based networks are trained to a fixed resolution. When the spatial resolution of the observation changes, the convolutional feature scales accordingly, which severely degrades the performance. 
Conversely, VRR world model is only concerned with the \textit{local} subset of state vector, and its performance is invariant to the overall resolution. 
The only limiting factor for VRR agent is the search algorithm's run-time, which is not problematic for $13\times 13$ Sokoban.

\begin{figure}[t]
  \centering
  \includegraphics[width=\linewidth]{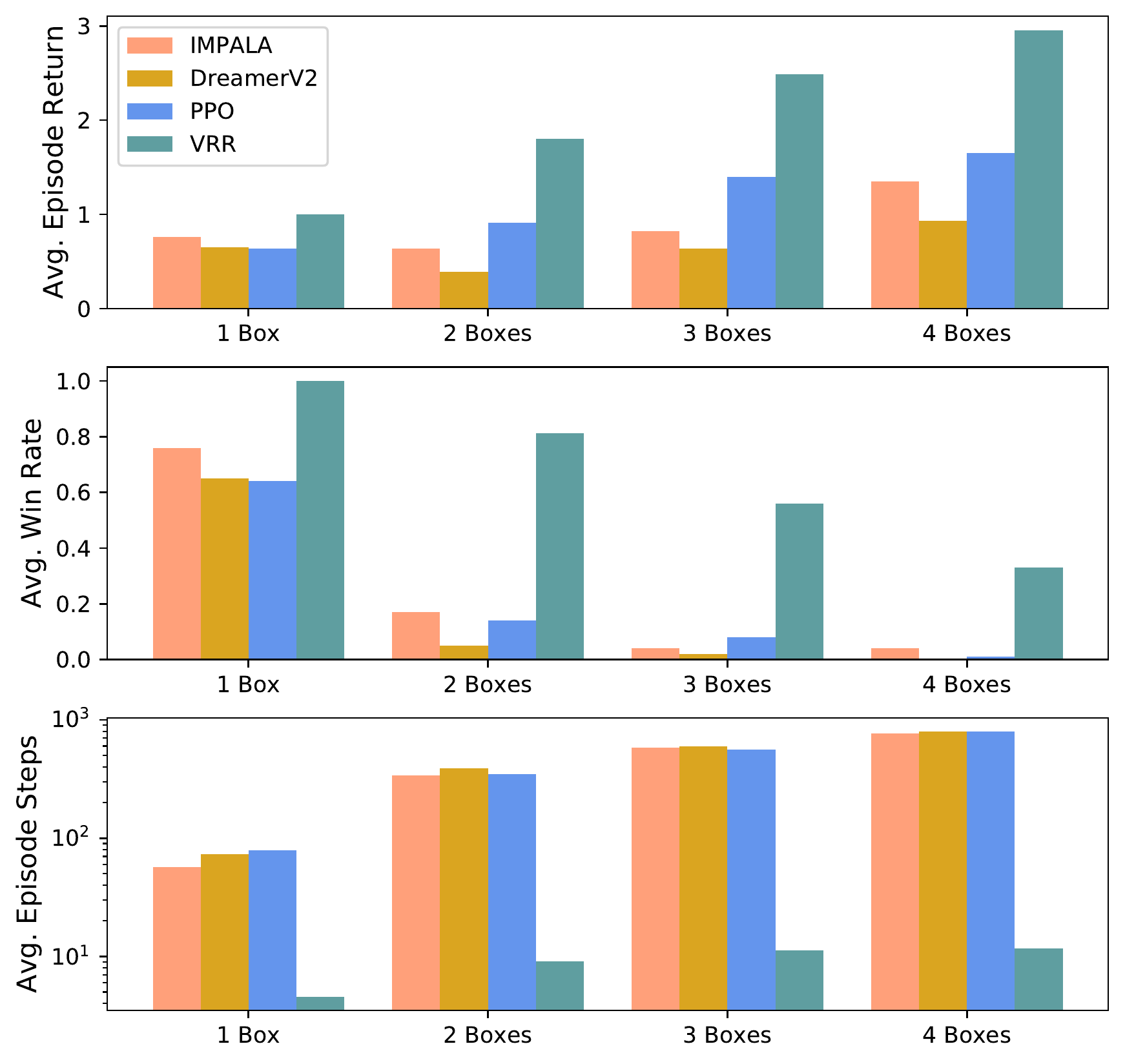}
  \caption{VRR and baseline agents on 7x7 Sokoban with varying number of boxes.}
  \label{fig:sokoban_box}
\end{figure}

We further test the agents' ability to solve in 2-, 3-, and 4-box Sokoban after being trained only on the 1-box Sokoban. Game board size is fixed to $7\times 7$.
Fig.~\ref{fig:sokoban_box} shows that VRR agent's performance degrades gracefully, and far outperforms baseline agents in the multiple-box Sokoban games. Note that the agent has never seen more than 1 box, and is disallowed to learn new rules. 
Notably, VRR agent not only accrues higher returns, but also solves each game episode with fewer steps. We accredit this advantage to the explicit planning via search tree in VRR agent. 
\begin{table}[H]
\centering
\begin{tabular}{c c c c c}
\hline
            & \multicolumn{4}{c}{Average Return}\\
Game        & VRR   & DreamerV2 & IMPALA & PPO\\
\hline
Original    & 1.0   &  0.91     & 1.0    &  1.0        \\
Rotated     & 1.0   &  0.07     & 0.43   &  0.37        \\
\hline
\end{tabular}
\caption{\label{tab:doorkey_rotate} Zero-shot agent performance on the randomly rotated DoorKey environment.}
\end{table}

\subsubsection{DoorKey}
First, the VRR and baseline agents are tested in a larger $32\times 32$ DoorKey environment (Fig.~\ref{fig:game_desc}, pic.(6)), where the layout is exactly identical: the agent starting position and the key are located to the left of the vertical wall, and the goal is located at the bottom right.   
Similar to Sokoban, Table~\ref{tab:grid_size} shows that VRR agent performance is invariant to board size, while baselines completely fail.

Additionally, we test VRR and baseline agents in $6\times 6$ DoorKey environment with randomly rotated initial states. For example, the goal state and agent initial positions may be swapped, or the agent may need to move upwards to open the door, instead of to the right.
Table~\ref{tab:doorkey_rotate} shows that VRR agent generalizes to rotated game board layouts, while baselines overfit to a particular orientation of the game board. The generalization ability of VRR agent is due to the VRR world model is invariant to rotations by design.



\section{Conclusion and Future Work}
In this paper, we attacked the generalization problem in RL using the concept of visual rewrite rules (VRR). VRR agents maintain a set of action-dependent graphical rules that describe action effects as local visual changes around the agent. Inspired by human game-playing priors, VRR models environment dynamics as minimal factored local changes. Given its simplicity and locality, a VRR agent can be trained with orders of magnitudes less data but still generalize better than the deep RL agents. Though we demonstrate the effectiveness of our method in grid-based environments, many open questions emerge from the assumptions of VRR. For example, can visual rewrite rules be easily generalized to continuous space environments? We suppose one could easily plug in an object detector to get the factored state observation and follow the same VRR algorithm, but we do not know if the training cost of the object detector will cancel out the efficiency and generalization ability of VRR. We also do not deny that deep representation learning is essential to high-dimensional problems. So it is natural to ask if the idea of visual rewrite is compatible with deep representation learning. Will this approach impose new inductive biases in deep agent design? Relaxations of any of the assumptions identified in the paper would be interesting future topics.

\bibliography{aaai22}

\end{document}